%% file: acl_latex.tex
\pdfoutput=1

\documentclass[11pt]{article}

\usepackage[final]{acl}

\usepackage{times}
\usepackage{latexsym}

\usepackage[T1]{fontenc}

\usepackage[utf8]{inputenc}

\usepackage{microtype}

\usepackage{inconsolata}

\usepackage{graphicx}

\usepackage{booktabs}
\usepackage{pifont}
\usepackage{colortbl} 
\usepackage{xcolor} 
\usepackage{subcaption}
\usepackage{xspace}
\usepackage{amsmath}

\newcommand{\langpair}[2]{\textsc{#1}$\rightarrow$\textsc{#2}}
\newcommand{\acceq}{acc$^{*}_{eq}$\xspace}

\newcommand{\cross}{\ding{55}}

\colorlet{lightyellow}{yellow!20}
\colorlet{lightred}{red!20}    
\colorlet{lightblue}{blue!20} 
\colorlet{lightgreen}{green!20} 
\colorlet{highlight}{gray!25}

\title{Has Machine Translation Evaluation Achieved Human Parity? \\ The Human Reference and the Limits of Progress}

\author{
  Lorenzo Proietti\textsuperscript{*} \quad
  Stefano Perrella\textsuperscript{*} \quad 
  {\bf Roberto Navigli} \\
  Sapienza NLP Group, Sapienza University of Rome \\
  \small\texttt{\{lproietti, perrella, navigli\}@diag.uniroma1.it}
}

\begin{document}
\maketitle

\def\thefootnote{*}\footnotetext{Equal contribution.}
\def\thefootnote{\arabic{footnote}}

\begin{abstract}

In Machine Translation (MT) evaluation, metric performance is assessed based on agreement with human judgments. In recent years, automatic metrics have demonstrated increasingly high levels of agreement with humans. To gain a clearer understanding of metric performance and establish an upper bound, we incorporate human baselines in the MT meta-evaluation, that is, the assessment of MT metrics' capabilities. 
Our results show that human annotators are not consistently superior to automatic metrics, with state-of-the-art metrics often ranking on par with or higher than human baselines. Despite these findings suggesting human parity, we discuss several reasons for caution. Finally, we explore the broader implications of our results for the research field, asking: Can we still reliably measure improvements in MT evaluation? 
With this work, we aim to shed light on the limits of our ability to measure progress in the field, fostering discussion on an issue that we believe is crucial to the entire MT evaluation community.

\end{abstract}

\section{Introduction and Related Work}

Machine Translation (MT) evaluation is the task of assessing the quality of translated text, while MT meta-evaluation estimates the capabilities of automatic evaluation techniques, i.e., MT metrics. 
Historically, automatic metrics have been employed due to their low cost and fast experimentation time, whereas human evaluation is still considered the gold standard, necessary for validating automatically-derived findings.
However, in recent years the MT evaluation field has seen significant advancements. Neural-based metrics have demonstrated strong correlations with human judgments, largely replacing traditional heuristic-based metrics, and becoming the de facto standard in MT evaluation \cite{freitag-etal-2022-results, freitag-etal-2023-results, freitag-etal-2024-llms}. More recently, LLM-based approaches to MT evaluation have emerged \cite{kocmi-federmann-2023-large, kocmi-federmann-2023-gemba, fernandes-etal-2023-devil, bavaresco2024llmsinsteadhumanjudges}, offering not only high correlation with human judgments but also improved interpretability. This raises the question of what is still missing in order for automatic techniques to achieve human parity, if they have not already.
Indeed, unlike other Natural Language Processing tasks, MT evaluation lacks a human performance reference, making it difficult to gauge the true capabilities of MT metrics. For instance, in MT, human performance is measured by evaluating human references alongside system translations \cite{laubli-etal-2018-machine, kocmi-etal-2023-findings, kocmi-etal-2024-findings}. Similarly, popular benchmarks such as HellaSwag \cite{zellers-etal-2019-hellaswag}, MMLU \cite{hendrycks2021measuring}, and MT-bench \cite{zheng2023judgingllmasajudgemtbenchchatbot} report the performance of human baselines. 

Since MT metrics' performance is measured based on agreement with human annotators, we posit that agreement among the annotators themselves can serve as a reference for human performance. Previous studies have reported the Inter-Annotator Agreement (IAA) in MT evaluation: \citet{pub7445} used Cohen’s kappa to measure the pairwise agreement between raters; \citet{freitag-etal-2021-experts} grouped raters' assessments into seven score bins before calculating pairwise agreement; and \citet{kocmi-etal-2024-error} used Kendall correlation coefficient $\tau_c$.
However, these studies employed different measures, making direct comparisons difficult, and none contextualized IAA in relation to the performance of automatic metrics. To the best of our knowledge, \citet{perrella-etal-2024-beyond} were the first to assess metric and human performance jointly. Specifically, they evaluated automatic metrics and human annotators within their new evaluation framework. Nonetheless, since comparing humans and metrics was not their primary focus, they included only one human annotation protocol  -- i.e., Direct Assessment + Scalar Quality Metrics \cite{kocmi-etal-2022-findings} -- that exhibited very poor performance, likely due to low annotation quality, rendering it ineffective as a human performance reference for MT metrics.

In this work, we address this gap by incorporating human baselines into the metric rankings from various editions of the Metrics Shared Task of the Conference on Machine Translation (WMT). By using meta-evaluation strategies from WMT 2024 we derive a single, comprehensive ranking of MT \textit{evaluators} -- both human and automatic -- establishing a human performance reference for MT metrics across several test sets, translation directions, and human annotation protocols, and offering a clearer understanding of the capabilities of current MT evaluation techniques. 
Then, given that our results suggest that automatic metrics may have reached human parity, we critically examine this claim and discuss its implications for future research in MT evaluation.\footnote{We publish the code to reproduce our results at \url{https://github.com/SapienzaNLP/human-parity-mt-eval}.}

\input{tables/human_evaluators_availability}

\input{tables/rankings}

\section{Preliminaries and Experimental Setup}
In this section, we describe the human annotations, the annotation protocols, the test sets selected for our work, the meta-evaluation strategies employed, and the automatic metrics included. 

\subsection{The Human Annotations} \label{sec:human_annotation}
Each year, WMT conducts manual annotation campaigns to collect human judgments of translation quality. First, each test set $t$ is created by drawing $N_t$ segments from various sources. Segments may consist of individual sentences or entire paragraphs. Each source segment is then translated into the target language using $M_t$ MT systems, producing $N_t \times M_t$ translations per test set $t$. Finally, human raters assess translation quality \cite{kocmi-etal-2023-findings, kocmi-etal-2024-findings, freitag-etal-2023-results, freitag-etal-2024-llms}. 

Given the large volume of translations, non-overlapping portions of each test set are typically assigned to different raters. Consequently, the annotated test sets used in this work combine annotations from multiple raters. For simplicity, we use the term \textit{evaluator} to refer to any entity that produced a set of annotations covering all segments in a test set. An evaluator can be a human rater, an MT metric, an ensemble of MT metrics, or an entity that selects annotations from different raters. For example, in the test set that we dub ``2020 \langpair{en}{de}'' \cite{freitag-etal-2021-experts}, six raters provided a total of three annotations per translation, yielding three distinct evaluators. 

However, this setup introduces a problem: Distinct human evaluators may be derived from non-disjoint sets of raters. If the same rater contributes to multiple evaluators, even across non-overlapping segments, it can artificially inflate their agreement and overestimate human baseline performance. To avoid this, we restrict each test set to the largest subset of segments annotated by strictly disjoint sets of raters. 
Returning to the 2020 \langpair{en}{de} example, we aim to partition the six raters into three groups, so that the combined annotations from raters within each group form a single evaluator.
Yet, two factors prevent such a simple partitioning: i) not all raters annotated every source segment, and ii) the specific rater-to-segment assignment prevents partitioning raters such that the combined annotations from each group cover all segments. Therefore, we restrict our test set to the segments that allow such a partitioning by solving the following optimization problem: \textit{Find the largest subset of segments and a partitioning of raters into three disjoint groups such that each group cumulatively annotated the entire subset of segments}. We apply a similar procedure to each test set with annotations of this form, reporting resulting test set sizes in Table~\ref{tab:employed_datasets}. Further details are provided in Appendix~\ref{apx:independence}. 

\subsection{Test Sets and Annotation Protocols} \label{sec:test-sets}
We estimate human performance based on the agreement among human evaluators. Specifically, we designate one human evaluator as ground truth while the others serve as human baselines. Consequently, our setup necessitates multiple human annotations for the same translations. Test sets satisfying this requirement include those released by \citet{freitag-etal-2021-experts} and those from WMT editions between 2022 and 2024.

These test sets feature human annotations from at least two of the following protocols: Multidimensional Quality Metrics (MQM, \citealp{ddd.uab.cat:130144}), Error Span Annotation (ESA, \citealp{kocmi-etal-2024-error}), Professional Scalar Quality Metrics (pSQM, \citealp{freitag-etal-2021-experts}), and Direct Assessments + Scalar Quality Metrics (DA+SQM, \citealp{kocmi-etal-2022-findings}). Our work leverages these test sets, but we restrict them to source segments that simultaneously: i) were annotated by all considered human evaluators and ii) were annotated by disjoint sets of raters (as detailed in Section~\ref{sec:human_annotation}).  
Table~\ref{tab:employed_datasets} presents statistics for the test sets employed. Additionally, we illustrate all the considered annotation protocols in Appendix~\ref{apx:protocols}.

Following standard practice in the literature \cite{freitag-etal-2021-experts, freitag-etal-2021-results, freitag-etal-2022-results, freitag-etal-2023-results, freitag-etal-2024-llms}, we designate evaluators derived from the MQM annotations released annually at WMT as the ground truth, employing the others as human baselines. Indeed, the MQM protocol relies on experienced annotators and provides a more detailed (and more expensive) evaluation compared to other protocols.  Nonetheless, in Appendix~\ref{apx:varying-ground-truth}, we also investigate the effects of selecting alternative evaluators -- either MQM evaluators different from those previously used or evaluators following different protocols -- as the ground truth.

\subsection{The MT Meta-Evaluation} \label{sec:meta-evaluation}
We compute metric rankings using the meta-evaluation strategies employed at the WMT 2024 Metrics Shared Task:
\begin{itemize}
    \item \textbf{Soft Pairwise Accuracy (SPA)} estimates evaluator performance based on the ability to rank \textit{MT systems}\footnote{The score assigned to an MT system is the average of the scores given to its translations.} in the same order as in the ranking derived from ground truth annotations \cite{thompson-etal-2024-improving}.

    \item \textbf{Pairwise Accuracy with Tie Calibration (\acceq)} estimates evaluator performance based on the ability to rank \textit{translations of the same source segment} in the same order as in the ranking derived from ground truth annotations \cite{deutsch-etal-2023-ties}.
\end{itemize}
We describe these measures in more detail in Appendix~\ref{apx:meta-evaluation}.

\subsection{Metrics}
The automatic evaluators considered -- i.e., the MT metrics -- are those submitted to the WMT Metrics Shared Task in the 2020, 2022, 2023, and 2024 editions. Additionally, we include several state-of-the-art metrics from recent WMT editions in rankings from previous years, provided they were not trained on the corresponding test sets. Table~\ref{tab:metrics} in Appendix~\ref{apx:metrics} lists all considered metrics. 

\section{Results}
Table~\ref{tab:rankings} presents the evaluator rankings. Due to space constraints, each table includes only a subset of evaluators. A complete set of results, including all the evaluators, is provided in Appendix~\ref{apx:full-rankings}. 

Results vary across years and translation directions. Notably, human evaluators do not consistently rank higher than automatic metrics. Under SPA, human evaluators often share clusters of statistical significance with automatic metrics, whereas, under \acceq, they are frequently surpassed. For example, in 2020 \langpair{en}{de}, BLEURT-0.2 and BLEURT-20 fall within the same statistical significance cluster as MQM-2020-3 and pSQM-2 under SPA, with pSQM-2 ranking as low as $9$th under \acceq. Similarly, in 2022 \langpair{en}{de}, MQM-2022-2 and MQM-2022-3 share the top cluster with MetricX-23-QE-XXL under SPA, with MQM-2022-2 ranking $6$th under \acceq. Finally, in 2023 and 2024, most human evaluators rank consistently below various automatic metrics under both SPA and \acceq. Even when restricted to the human evaluators who follow the same protocol as the annotations employed as gold -- i.e., MQM -- they rank consistently in the top positions solely in 2020. Additionally, our findings remain valid when varying the human evaluators used as ground truth, as shown in Appendix~\ref{apx:varying-ground-truth}.

These results may indicate human-level performance in MT evaluation. Nonetheless, we argue that they are insufficient to establish equivalence between human and automatic evaluators, and elaborate our reasons in the next section.  

\section{Discussion}
In the same spirit as \citet{tedeschi-etal-2023-whats}, who discuss the meaning of superhuman performance in Natural Language Understanding, we outline several factors to consider before making similar claims in MT evaluation. We then discuss the broader implications of our findings, warning that measuring progress in the field may become increasingly challenging.

\paragraph{Meta-evaluation}
Certain meta-evaluation measures may be inadequate for comparing human and automatic evaluators. In particular, our results consistently rank human evaluators much lower under \acceq than under SPA. This discrepancy may be related to the findings of \citet{perrella-etal-2024-guardians}, who show that \acceq favors evaluators whose assessments fall within a continuous interval, whereas, as detailed in Appendix~\ref{apx:protocols}, human evaluators produce discrete assessments. 

\paragraph{Annotation quality}
Certain annotation campaigns might have produced low-quality annotations, either due to a lack of clarity in the annotation guidelines or to the ability of the raters involved. This is particularly concerning in the 2023 \langpair{en}{de} test set, where, even if restricted to SPA, most human evaluators fall within the second cluster of statistical significance, alongside surface-level metrics such as BLEU.\footnote{We wish to highlight that our 2023 test set features only $145$ segments annotated by all human evaluators (as reported in Table~\ref{tab:employed_datasets}), which might have resulted in unreliable estimates of SPA and \acceq.}

\paragraph{Benchmarks difficulty}
Current test sets might be too easy for the MT systems whose translations are being evaluated. Supporting this hypothesis, we observe that sentinel-cand-mqm, a metric that assesses only translation fluency, ranks on par with the human evaluator ESA under SPA, and even higher under \acceq (Table~\ref{tab:full-wmt24}). This suggests that the evaluated translations may differ only in minor fluency-related nuances. Arguably, to assess whether human parity has been truly achieved, future studies should compare metrics and humans in more demanding contexts. Indeed, previous research has shown that metrics may struggle in unseen domains \cite{zouhar-etal-2024-fine} and lack sensitivity to specific translation errors such as incorrect number, gender \cite{karpinska-etal-2022-demetr}, or word sense disambiguation \cite{10.1162/coli_a_00541}. 

\subsection{Can We Still Measure Improvements in MT Evaluation?}
As discussed, we believe claiming human parity is premature without first addressing the issues outlined above. Nonetheless, with automatic metrics ranking the same as, or higher than,  human evaluators in standard benchmarks, our results raise a critical concern about our ability to measure progress in MT evaluation: What does a higher or lower ranking truly mean? 

If a metric ranks higher than a human evaluator using a non-MQM protocol, is the metric a better evaluator, or does it merely align more closely with the score distribution of the MQM protocol? More concerningly, if a metric ranks higher than an MQM evaluator, does this suggest superior evaluation capabilities, or does it simply reflect better alignment with the specific raters who produced the gold annotations? Indeed, \citet{finkelstein2024jacktradesmasterone} achieved an exceptionally high agreement with gold annotations by explicitly optimizing their metric to align with the raters themselves. More generally, we argue that in current benchmarks it is unclear whether a higher ranking -- relative to either a human or an automatic evaluator -- reflects genuine improvements in evaluation quality or merely closer alignment with a particular annotation protocol or rater style.

To ensure the reliability of meta-evaluation, future research should focus on exploring whether the gap between human and automatic evaluators can be restored. This could be pursued in several ways, including (but not limited to) selecting more challenging test sets, using test sets adversarial to MT metrics (e.g., from domains different from their training data), producing higher-quality human annotations, or designing new annotation protocols that yield stronger inter-annotator agreement. Additionally, greater resources could be allocated to human annotation campaigns -- either by collecting multiple annotations per translation to reach a consensus among annotators or by increasing the number of segments in test sets, as suggested by \citet{riley-etal-2024-finding}.

\section{Conclusions}
We incorporate human baselines into the metric rankings from previous editions of the WMT Metrics Shared Task. Our results show that MT metrics frequently rank higher than human evaluators, particularly when the latter follow annotation protocols different from MQM -- the protocol used as the ground truth. While our findings suggest that metrics may have reached human-level performance, we recommend caution and highlight several issues the research community should address to assess whether human parity has been truly achieved. Finally, we discuss a critical concern arising from our findings: the limits of measuring progress in MT evaluation as automatic metrics approach human baselines. In this respect, we propose research directions to ensure that progress remains measurable or, at the very least, to extend the period during which it can be reliably tracked.

\clearpage

\section*{Limitations}
This study required test sets annotated by multiple human evaluators. Consequently, our analysis is limited to seven test sets including four language directions.

Moreover, assessing the agreement between various human evaluators required restricting our analysis to segments annotated by all of them. As a result, some test sets contain only a small number of segments, which might reduce the reliability of the results. To mitigate this issue, our findings are supported by statistical significance analyses. 


\section*{Acknowledgements}

\begin{center}
\noindent
    \begin{minipage}{0.1\linewidth}
        \begin{center}
            \includegraphics[scale=0.05]{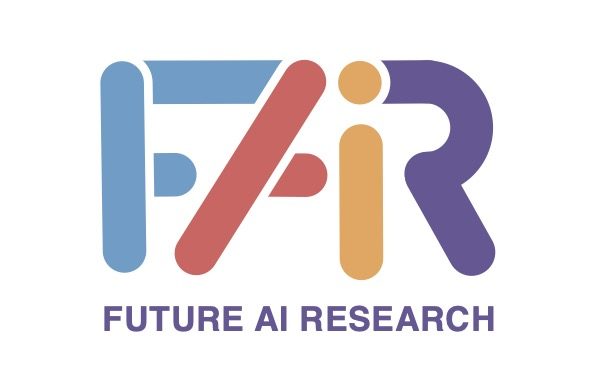}
        \end{center}
    \end{minipage}
    \hspace{0.01\linewidth}
    \begin{minipage}{0.70\linewidth}
         We gratefully acknowledge the support of the PNRR MUR project PE0000013-FAIR, and the CREATIVE project (CRoss-modal understanding and gEnerATIon of Visual and tExtual content), which is funded by the MUR Progetti di Rilevante Interesse Nazionale programme (PRIN 2020). The authors acknowledge the CINECA award IsCb9\_mtmit under the
         ISCRA initiative for the availability of high-performance
         computing resources.
    \end{minipage}
    \hspace{0.01\linewidth}
    \begin{minipage}{0.1\linewidth}
        \begin{center}
            \includegraphics[scale=0.08]{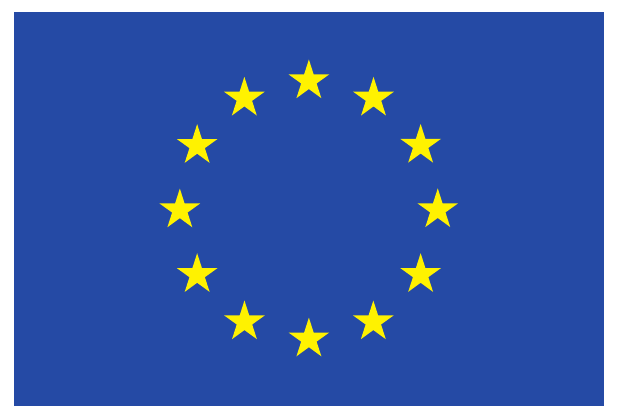}
        \end{center}
    \end{minipage}\\
\end{center}
\vspace{0.2cm}

\noindent This work was carried out while Lorenzo Proietti was enrolled in the Italian National Doctorate on Artificial Intelligence run by Sapienza University of Rome. The authors thank Anna Bavaresco for her valuable feedback on the manuscript.

\bibliography{anthology,custom}

\appendix

\input{latex/appendix}

\end{document}

%% file: tables/human_evaluators_availability.tex
\begin{table}[t]
    \centering
    \resizebox{0.99\linewidth}{!}{
    \begin{tabular}{l|c|c|c|c|c|c|c}
        \toprule
        &  \multicolumn{2}{c|}{2020}  & \multicolumn{2}{c|}{2022} & \multicolumn{2}{c|}{2023} & \multicolumn{1}{c}{2024} \\
        
        & \multicolumn{1}{c|}{\small $\rightarrow$ \textsc{de}} & \multicolumn{1}{c|}{\small \textsc{zh} $\rightarrow$ } & \multicolumn{1}{c|}{\small $\rightarrow$ \textsc{de}} & \multicolumn{1}{c|}{\small $\rightarrow$ \textsc{zh}} & \multicolumn{1}{c|}{\small $\rightarrow$ \textsc{de}} & \multicolumn{1}{c|}{\small \textsc{zh} $\rightarrow$} & \multicolumn{1}{c}{\small $\rightarrow$ \textsc{es}} \\
        \midrule

        MQM & $3$ & $3$ & $3$ & $3$ & $4$ & $3$ & $1$  \\
        ESA & \cross & \cross & \cross & \cross & $2$ & \cross & $1$ \\
        pSQM & $3$ & $3$ & \cross & \cross & \cross & \cross & \cross \\
        DA+SQM & \cross & \cross & $1$ & $1$ & $1$ & $1$ & \cross \\
        \midrule

        \#Seg & $681$ & $895$ & $583$ & $1065$ & $145$ & $687$ & $449$ \\
        \#Sys & $9$ & $9$ & $10$ & $13$ & $12$ & $15$ & $12$ \\
    \end{tabular}
    }
    \caption{The four top rows indicate the number of distinct evaluators for each annotation protocol and test set. We list the studies that released these annotations in Appendix~\ref{apx:protocols}. `2020' refers to the data released by \citet{freitag-etal-2021-experts}, while other years correspond to the test sets from the corresponding WMT editions. The notation $\rightarrow$~\textsc{x} indicates that the test set contains translations from English to \textsc{x}, whereas \textsc{x}~$\rightarrow$ denotes translations from \textsc{x} to English. The two bottom rows present the number of source segments and automatic translations per source segment present in the intersection of annotations from all human evaluators, restricted to the segments annotated by disjoint sets of raters (§\ref{sec:human_annotation}).}
    \label{tab:employed_datasets}
\end{table}

%% file: tables/rankings.tex
\begin{table*}[ht!]
    \centering

    \resizebox{0.90\textwidth}{!}{
    \begin{subtable}{\textwidth}
        \centering

        \begin{tabular}{lrr|rr|rr|rr}
        \toprule
        \textbf{Test set $\mathbf{2020}$} & \multicolumn{4}{c|}{\langpair{en}{de}}& \multicolumn{4}{c}{\langpair{zh}{en}}\\
        & \multicolumn{2}{c|}{SPA}& \multicolumn{2}{c|}{\acceq}& \multicolumn{2}{c|}{SPA}& \multicolumn{2}{c}{\acceq}\\
        \multicolumn{1}{l}{Metric} & \multicolumn{1}{c}{ Rank } & \multicolumn{1}{c|}{ Acc. } & \multicolumn{1}{c}{ Rank } & \multicolumn{1}{c|}{ Acc. } & \multicolumn{1}{c}{ Rank } & \multicolumn{1}{c|}{ Acc. } & \multicolumn{1}{c}{ Rank } & \multicolumn{1}{c}{ Acc. }\\
        \midrule
        \cellcolor{highlight}MQM-2020-2& \cellcolor{highlight} $1$ & \cellcolor{highlight} $96.45$& \cellcolor{highlight} $1$ & \cellcolor{highlight} $58.86$& \cellcolor{highlight} $1$ & \cellcolor{highlight} $88.10$& \cellcolor{highlight} $1$ & \cellcolor{highlight} $55.70$\\
        \cellcolor{highlight}pSQM-1& \cellcolor{highlight} $1$ & \cellcolor{highlight} $95.59$& \cellcolor{highlight} $6$ & \cellcolor{highlight} $49.41$& \cellcolor{highlight} $1$ & \cellcolor{highlight} $79.16$& \cellcolor{highlight} $13$ & \cellcolor{highlight} $43.89$\\
        \cellcolor{highlight}MQM-2020-3& \cellcolor{highlight} $2$ & \cellcolor{highlight} $90.39$& \cellcolor{highlight} $2$ & \cellcolor{highlight} $56.84$& \cellcolor{highlight} $1$ & \cellcolor{highlight} $92.06$& \cellcolor{highlight} $2$ & \cellcolor{highlight} $52.80$\\
        BLEURT-0.2& $2$ & $86.81$& $4$ & $50.81$& $2$ & $72.59$& $3$ & $50.57$\\
        \cellcolor{highlight}pSQM-2& \cellcolor{highlight} $2$ & \cellcolor{highlight} $85.87$& \cellcolor{highlight} $9$ & \cellcolor{highlight} $46.97$& \cellcolor{highlight} $1$ & \cellcolor{highlight} $89.33$& \cellcolor{highlight} $9$ & \cellcolor{highlight} $46.77$\\
        BLEURT-20& $2$ & $85.52$& $3$ & $51.68$& $3$ & $67.46$& $4$ & $50.12$\\
        \bottomrule \\ [-1.5ex]
        
        \toprule
        \textbf{Test set $\mathbf{2022}$} & \multicolumn{4}{c|}{\langpair{en}{de}}& \multicolumn{4}{c}{\langpair{en}{zh}}\\
        & \multicolumn{2}{c|}{SPA}& \multicolumn{2}{c|}{\acceq}& \multicolumn{2}{c|}{SPA}& \multicolumn{2}{c}{\acceq}\\
        \multicolumn{1}{l}{Metric} & \multicolumn{1}{c}{ Rank } & \multicolumn{1}{c|}{ Acc. } & \multicolumn{1}{c}{ Rank } & \multicolumn{1}{c|}{ Acc. } & \multicolumn{1}{c}{ Rank } & \multicolumn{1}{c|}{ Acc. } & \multicolumn{1}{c}{ Rank } & \multicolumn{1}{c}{ Acc. }\\
        \midrule
        MetricX-23-QE-XXL*& $1$ & $94.89$& $3$ & $57.64$& $2$ & $83.92$& $2$ & $47.43$\\
        \cellcolor{highlight}MQM-2022-2& \cellcolor{highlight} $1$ & \cellcolor{highlight} $94.49$& \cellcolor{highlight} $6$ & \cellcolor{highlight} $55.55$& \cellcolor{highlight} $2$ & \cellcolor{highlight} $80.82$& \cellcolor{highlight} $3$ & \cellcolor{highlight} $47.05$\\
        \cellcolor{highlight}MQM-2022-3& \cellcolor{highlight} $1$ & \cellcolor{highlight} $92.59$& \cellcolor{highlight} $1$ & \cellcolor{highlight} $61.06$& \cellcolor{highlight} $1$ & \cellcolor{highlight} $87.22$& \cellcolor{highlight} $2$ & \cellcolor{highlight} $47.56$\\
        MetricX-23-XXL& $2$ & $92.34$& $2$ & $59.27$& $1$ & $87.69$& $1$ & $48.43$\\
        \cellcolor{highlight}DA+SQM& \cellcolor{highlight} $6$ & \cellcolor{highlight} $66.61$& \cellcolor{highlight} $16$ & \cellcolor{highlight} $46.03$& \cellcolor{highlight} $2$ & \cellcolor{highlight} $82.95$& \cellcolor{highlight} $12$ & \cellcolor{highlight} $36.26$\\
        \bottomrule \\ [-1.5ex]
        
        \toprule
        \textbf{Test set $\mathbf{2023}$} & \multicolumn{4}{c|}{\langpair{en}{de}} & \multicolumn{4}{c}{\langpair{zh}{en}}  \\
        & \multicolumn{2}{c|}{SPA}& \multicolumn{2}{c|}{\acceq} & \multicolumn{2}{c|}{SPA}& \multicolumn{2}{c}{\acceq}\\
        \multicolumn{1}{l}{Metric} & \multicolumn{1}{c}{ Rank } & \multicolumn{1}{c|}{ Acc. } & \multicolumn{1}{c}{ Rank } & \multicolumn{1}{c|}{ Acc. } & \multicolumn{1}{c}{ Rank } & \multicolumn{1}{c|}{ Acc. } & \multicolumn{1}{c}{ Rank } & \multicolumn{1}{c}{ Acc. }\\
        \midrule
        GEMBA-MQM*& $1$ & $94.52$& $5$ & $58.52$ & $1$ & $93.17$ & $3$ & $52.80$\\
        \cellcolor{highlight}MQM-2023-3& \cellcolor{highlight} $1$ & \cellcolor{highlight} $93.51$& \cellcolor{highlight} $5$ & \cellcolor{highlight} $58.42$ & \cellcolor{highlight} $1$ & \cellcolor{highlight} $95.54$& \cellcolor{highlight} $5$ & \cellcolor{highlight} $51.65$ \\
        \cellcolor{highlight}MQM-2023-2& \cellcolor{highlight} $1$ & \cellcolor{highlight} $93.15$& \cellcolor{highlight} $6$ & \cellcolor{highlight} $57.71$ & \cellcolor{highlight} $1$ & \cellcolor{highlight} $95.18$& \cellcolor{highlight} $2$ & \cellcolor{highlight} $52.90$ \\
        XCOMET-Ensemble & $1$ & $92.21$ & $3$ & $60.99$ & $2$ & $91.15$ & $1$ & $54.59$ \\
        MetricX-23-QE-XXL* & $1$ & $92.12$& $1$ & $62.53$  & $3$ & $88.30$& $2$ & $53.26$ \\
        \cellcolor{highlight}DA+SQM & \cellcolor{highlight} $2$ & \cellcolor{highlight} $91.24$& \cellcolor{highlight} $14$ & \cellcolor{highlight} $46.79$ & \cellcolor{highlight} $4$ & \cellcolor{highlight} $86.28$& \cellcolor{highlight} $22$ & \cellcolor{highlight} $39.42$ \\
        \cellcolor{highlight}ESA-1& \cellcolor{highlight} $2$ & \cellcolor{highlight} $90.39$& \cellcolor{highlight} $14$ & \cellcolor{highlight} $46.71$ & \cellcolor{highlight}-- &\cellcolor{highlight} -- &\cellcolor{highlight} -- &\cellcolor{highlight} -- \\
        \cellcolor{highlight}ESA-2& \cellcolor{highlight} $2$ & \cellcolor{highlight} $89.11$& \cellcolor{highlight} $12$ & \cellcolor{highlight} $49.70$ & \cellcolor{highlight}-- & \cellcolor{highlight}-- & \cellcolor{highlight}-- &\cellcolor{highlight} --\\
        \cellcolor{highlight}MQM-2023-4& \cellcolor{highlight} $2$ & \cellcolor{highlight} $88.93$& \cellcolor{highlight} $14$ & \cellcolor{highlight} $46.68$ & \cellcolor{highlight} -- & \cellcolor{highlight} -- & \cellcolor{highlight} -- & \cellcolor{highlight} -- \\
        \bottomrule
        \end{tabular}
        
    \end{subtable}
    }

    \vspace{0.2cm}

    \resizebox{0.90\textwidth}{!}{
    \begin{subtable}{\textwidth}
        \centering
        \begin{tabular}{lrr|rr}
        \toprule
        \textbf{Test set $\mathbf{2024}$} & \multicolumn{4}{c}{\langpair{en}{es}} \\
        & \multicolumn{2}{c|}{SPA}& \multicolumn{2}{c}{\acceq}\\
        \multicolumn{1}{l}{Metric} & \multicolumn{1}{c}{ Rank } & \multicolumn{1}{c|}{ Acc. } & \multicolumn{1}{c}{ Rank } & \multicolumn{1}{c}{ Acc. }\\
        \midrule
        CometKiwi-XXL*& $1$ & $86.12$& $4$ & $67.24$\\
        gemba\_esa*& $1$ & $85.72$& $3$ & $67.68$\\
        \cellcolor{highlight}ESA& \cellcolor{highlight} $2$ & \cellcolor{highlight} $80.12$& \cellcolor{highlight} $8$ & \cellcolor{highlight} $63.84$\\
        metametrics\_mt\_mqm\_hybrid\_kendall & $2$ & $80.10$& $1$ & $68.95$\\
        MetricX-24-Hybrid& $2$ & $79.75$& $1$ & $69.20$\\
        \bottomrule
        \end{tabular}

        \label{tab:wmt2024}
        
    \end{subtable}
    }
    
    \caption{Results obtained by applying the WMT 2024 Meta-Evaluation strategies to the test sets illustrated in Section~\ref{sec:test-sets}. The `Acc.' column contains the Meta-Evaluation accuracy, while `Rank' reports clusters of statistical significance computed following \citet{freitag-etal-2024-llms}, using the PERM-BOTH hypothesis test introduced by \citet{deutsch-etal-2021-statistical}. Starred metrics are reference-less metrics, and rows highlighted in gray are human evaluators.}
    \label{tab:rankings}
    
\end{table*}

%% file: latex/appendix.tex
\section{Fair Extraction of Evaluators from Human Annotations} \label{apx:independence}
The human evaluation campaigns conducted by \citet{freitag-etal-2021-experts}, \citet{freitag-etal-2023-results}, and \citet{riley-etal-2024-finding} produced multiple annotations for each translation. As you can see in Table~\ref{tab:employed_datasets}, there are many annotations per translation for MQM and pSQM in the test sets 2020, 2022, and 2023. As discussed in Section~\ref{sec:human_annotation}, these annotation campaigns distributed the annotation workload among multiple raters. 

Since we derive multiple evaluators from these annotations (some used as ground truth and some as human baselines), we prevent artificially inflating their agreement by not allowing the same rater to contribute to two distinct evaluators simultaneously. For example, in the 2020 \langpair{en}{de} test set, six raters provided a total of three annotations per translation. We extract three human evaluators from these annotations, using one as the ground truth and the other two as human evaluators (MQM-2020-2 and MQM-2020-3 in Table~\ref{tab:rankings}). To achieve this, we partition the six raters into three groups, each forming one evaluator. However, not all raters annotated the entire set of source segments, and the distribution of workload did not allow for a partition that covered all annotated segments. Therefore, to retain the maximum number of segments in our test sets, we solved the following optimization problem: \textit{Find the largest subset of segments and a partitioning of raters into three disjoint groups such that each group cumulatively annotated the entire subset of segments}.

Formally, let us define a test set $t = \{s_1, ..., s_{N_t}\}$ as a set of $N_t$ segments. Each segment was annotated by $k$ out of $R$ raters, with $\mathcal{R}=\{r_1, ..., r_R \}$ representing the set of raters. Our goal is to determine a partition $\Pi = \{\mathcal{R}_1, ..., \mathcal{R}_k\}$ of $\mathcal{R}$ and a subset $u \subseteq t$ such that $u$ is the largest subset in which every segment has been annotated by exactly one rater from each of the $k$ sets in the partition $\Pi$. 

To solve this optimization problem, we formulate it as an Integer Linear Programming (ILP) problem and solve it using the PuLP\footnote{\url{https://coin-or.github.io/pulp/}.} Python library. We applied this procedure to the 2020, 2022, and 2023 test sets.

\section{Human Annotations} \label{apx:protocols}
We briefly illustrate how each annotation protocol considered works:
\begin{itemize}
    \item Multidimensional Quality Metrics (MQM) requires annotators to identify error spans in the translated text, specifying error category and severity, to be selected among Neutral, Minor, Major, and Critical. A translation quality score is derived by assigning a penalty to each error span depending on severity \cite{ddd.uab.cat:130144,freitag-etal-2021-experts}.
    
    \item Error Span Annotation (ESA) requires annotators to identify error spans in the translated text, specify error severity, and later assign a scalar quality score from 0 to 100 to the translation \cite{kocmi-etal-2024-error}.

    \item Scalar Quality Metrics (SQM) requires annotators to assign a scalar quality score from 0 to 6 to the translated text. Following \cite{freitag-etal-2021-experts}, we use `pSQM' to refer to SQM conducted by professional annotators.\footnote{In this work, we use only annotations produced by professional annotators or translators. Therefore, we exclude cSQM and Direct Assessments (DA) -- which were crowdsourced -- from the 2020 test sets.} 

    \item Direct Assessments + Scalar Quality Metrics \cite[DA+SQM]{kocmi-etal-2022-findings} requires raters to assign a scalar quality score from 0 to 100 to the translated text. Raters are presented with seven labeled tick marks describing translation quality levels at various score thresholds, similarly to the SQM protocol.
\end{itemize}

Here, for each set of annotations employed in this work (i.e., those reported in Table~\ref{tab:employed_datasets}), we indicate the reference paper that released them:
\begin{itemize}
    \item The MQM-based and pSQM-based annotations for the test sets 2020 \langpair{en}{de} and 2020 \langpair{zh}{en} have been released by \citet{freitag-etal-2021-experts}.

    \item The MQM-based annotations for the test sets 2022 \langpair{en}{de} and 2022 \langpair{en}{zh} have been released by \citet{freitag-etal-2022-results} and \citet{riley-etal-2024-finding}. 

    \item The DA+SQM-based annotations for the test sets 2022 \langpair{en}{de} and 2022 \langpair{en}{zh} have been released by \citet{kocmi-etal-2022-findings}.

    \item Three sets of MQM-based annotations for the test sets 2023 \langpair{en}{de} and \langpair{zh}{en} have been released by \citet{freitag-etal-2023-results}.

    \item The ESA-based annotations and the last set of MQM-based annotations (MQM-2023-4 in Table~\ref{tab:rankings}) for the test set 2023 \langpair{en}{de} have been released by \citet{kocmi-etal-2024-error}.

    \item The ESA-based annotations for the test set 2024 \langpair{en}{es} have been released by \citet{kocmi-etal-2024-findings}.

    \item The MQM-based annotations for the test set 2024 \langpair{en}{es} have been released by \citet{freitag-etal-2024-llms}.
\end{itemize}

\section{Meta-Evaluation Measures} \label{apx:meta-evaluation}
In this section, we describe the two meta-evaluation measures used in our work, as listed in Section~\ref{sec:meta-evaluation}.

\subsection{Soft Pairwise Accuracy (SPA)}
\citet{thompson-etal-2024-improving} introduced Soft Pairwise Accuracy (SPA) as an extension of Pairwise Accuracy \cite[PA]{kocmi-etal-2021-ship}.

Given a test set $t$, which consists of $N_t$ source segments and $M_t$ translations generated by the respective $M_t$ MT systems (as described in Section~\ref{sec:human_annotation}), PA counts how often an evaluator $e$ ranks system pairs in the same order as the ground truth $g$. Let $a_{ij}$ be 1 if evaluator $e$ ranks systems $i$ and $j$ in the same order as the ground truth and $0$ otherwise, where $i,j \in \{0, ..., Mt\}$. Then, PA is defined as:
\begin{equation}
    PA = \binom{N}{2}^{-1} \sum_{i=0}^{M_t} \sum_{j=i+1}^{M_t} a_{ij}
\end{equation}

SPA extends PA by incorporating the confidence with which an evaluator and the ground truth rank two MT systems. Confidence is represented using statistical $p$-values. Specifically, $p_{ij}^{e}$ denotes the $p$-value associated with the statistical hypothesis that system $i$ is better than system $j$ according to evaluator $e$, while $p_{ij}^g$ represents the corresponding $p$-value for the ground truth $g$. SPA is then defined as follows:
\begin{equation}
    SPA = \binom{N}{2}^{-1} \sum_{i=0}^{M_t} \sum_{j=i+1}^{M_t} 1 - |p_{ij}^g - p_{ij}^e |
\end{equation}
Thus, SPA rewards an evaluator for expressing confidence levels similar to those of the ground truth and penalizes deviations.

\subsection{Pairwise Accuracy with Tie Calibration (\texorpdfstring{\acceq}{acceq})}

\citet{deutsch-etal-2023-ties} introduced \acceq to account for tied scores in meta-evaluation. Unlike PA and SPA, \acceq is a segment-level measure, meaning it evaluates a metric's ability to estimate the quality of individual translations rather than MT systems. Specifically, \acceq counts how often an evaluator $e$ ranks pairs of translations of the same source segment in the same order as the ground truth $g$, accounting for tied scores.

Let $C$ be the number of translation pairs ranked in the same order by both the evaluator $e$ and the ground truth $g$. Similarly, let $D$ denote the pairs ranked in the opposite order. The terms $T_e$ and $T_g$ represent pairs tied only in the evaluator's scores and only in the ground truth, respectively. Lastly, $T_{eg}$ refers to pairs tied in both the evaluator's scores and the ground truth. \acceq is then defined as:

\begin{equation}
    \text{\acceq} = \frac{C + T_{eg}}{C + D + T_e + T_g + T_{eg}}
\end{equation}

\paragraph{Tie Calibration}
Many automatic metrics produce assessments on a continuous scale, such as the real numbers in the interval $[0,1]$. As a consequence, these metrics rarely, if ever, produce tied scores, resulting in $T_e \approx 0$ and $T_{eg} \approx 0$. The Tie Calibration algorithm addresses this issue by estimating a threshold value  $\epsilon_e$ for each evaluator $e$, such that two assessments $e_i$ and $e_j$ are considered tied if $|e_i - e_j| \leq \epsilon_e$.

\section{Metrics} \label{apx:metrics}
Table~\ref{tab:metrics} lists the complete set of automatic evaluators considered in this work.

\input{tables/metrics}

\section{Full Rankings} \label{apx:full-rankings}
Tables~\ref{tab:full-wmt2020}, \ref{tab:full-wmt2022}, \ref{tab:full-wmt2023}, and \ref{tab:full-wmt24} present the same rankings of Table~\ref{tab:rankings}, but including all tested evaluators.

\input{tables/wmt2020}
\input{tables/wmt2022}
\input{tables/wmt2023}
\input{tables/wmt2024}

\section{Full Rankings Varying the Ground Truth} \label{apx:varying-ground-truth}

In this section, we examine how evaluator rankings vary depending on the choice of human evaluator used as ground truth. Specifically, we use the following evaluators as ground truth:
\begin{itemize}
    \item pSQM-1 from Table~\ref{tab:full-wmt2020}.
    \item DA+SQM from Table~\ref{tab:full-wmt2023}.
    \item MQM-2023-2 from Table~\ref{tab:full-wmt2023}.
    \item MQM-2023-3 from Table~\ref{tab:full-wmt2023}.
    \item ESA from Table~\ref{tab:full-wmt24}.
\end{itemize}
We exclude the ESA-1, ESA-2, and MQM-2023-4 evaluators from the 2023 \langpair{en}{de} test set, as they annotated only a limited number of segments. This increases the number of segments in the 2023 \langpair{en}{de} test set from $145$ to $376$. Therefore, for reference, we also report results on this test set using the same evaluator as in Tables~\ref{tab:rankings} and \ref{tab:full-wmt2023}. Results are presented in Tables~\ref{tab:wmt20-gold-psqm}, \ref{tab:wmt23-gold-dasqm}, \ref{tab:wmt23-gold-mqm1}, \ref{tab:wmt23-gold-mqm2}, \ref{tab:wmt23-gold-mqm3}, and \ref{tab:wmt24-gold-esa}.

As we can see, our findings remain valid when varying the evaluator selected as ground truth, with human evaluators consistently ranking the same as or lower than automatic metrics.

\input{tables/varying_ground_truth/wmt2020_gold_psqm}
\input{tables/varying_ground_truth/wmt2023_gold_dasqm}
\input{tables/varying_ground_truth/wmt2023_gold_mqm1}
\input{tables/varying_ground_truth/wmt2023_gold_mqm2}
\input{tables/varying_ground_truth/wmt2023_gold_mqm3}

\input{tables/varying_ground_truth/wmt2024_gold_esa}

%% file: tables/metrics.tex
\begin{table*}[t]
\centering
\resizebox{\textwidth}{!}{
\begin{tabular}{llll}
\toprule
Metric & Reference paper & Metric & Reference paper \\
\midrule
all-rembert-20 & \cite{mathur-etal-2020-results} & metametrics\_mt\_mqm & \cite{anugraha-etal-2024-metametrics} \\
BAQ\_dyn & \cite{mathur-etal-2020-results} & metametrics\_mt\_mqm\_qe & \cite{anugraha-etal-2024-metametrics} \\
BAQ\_static & \cite{mathur-etal-2020-results} & MetricX-23-QE-XXL & \cite{juraska-etal-2023-metricx} \\
BERT-base-L2 & \cite{mathur-etal-2020-results} & MetricX-23-XXL & \cite{juraska-etal-2023-metricx} \\
BERT-large-L2 & \cite{mathur-etal-2020-results} & MetricX-24-Hybrid & \cite{juraska-etal-2024-metricx} \\
BERTScore & \cite{Zhang*2020BERTScore:} & MetricX-24-Hybrid-QE & \cite{juraska-etal-2024-metricx} \\
BLCOM\_1 & \cite{freitag-etal-2024-llms} & metricx\_xxl\_MQM\_2020 & \cite{freitag-etal-2022-results} \\
BLEU & \cite{papineni-etal-2002-bleu} & mre-score-labse-regular & \cite{viskov-etal-2023-semantically} \\
BLEURT & \cite{sellam-etal-2020-bleurt} & MS-COMET-22 & \cite{kocmi-etal-2022-ms} \\
BLEURT-0.1-all & \cite{mathur-etal-2020-results} & MS-COMET-QE-22 & \cite{kocmi-etal-2022-ms} \\
BLEURT-0.1-en & \cite{mathur-etal-2020-results} & OpenKiwi-Bert & \cite{kepler-etal-2019-openkiwi} \\
BLEURT-0.2 & \cite{mathur-etal-2020-results} & OpenKiwi-XLMR & \cite{kepler-etal-2019-openkiwi} \\
BLEURT-20 & \cite{sellam-etal-2020-bleurt} & parbleu & \cite{bawden-etal-2020-parbleu} \\
bleurt-combi & \cite{mathur-etal-2020-results} & parchrf++ & \cite{bawden-etal-2020-parbleu} \\
BLEURT-extended & \cite{sellam-etal-2020-learning} & paresim-1 & \cite{bawden-etal-2020-parbleu} \\
bright-qe & \cite{freitag-etal-2024-llms} & prism & \cite{thompson-post-2020-automatic} \\
Calibri-COMET22 & \cite{freitag-etal-2023-results} & prismRef & \cite{thompson-post-2020-automatic, thompson-post-2020-paraphrase} \\
Calibri-COMET22-QE & \cite{freitag-etal-2023-results} & PrismRefMedium & \cite{thompson-post-2020-automatic, thompson-post-2020-paraphrase} \\
CharacTER & \cite{wang-etal-2016-character} & PrismRefSmall & \cite{thompson-post-2020-automatic, thompson-post-2020-paraphrase} \\
chrF & \cite{popovic-2015-chrf} & prismSrc & \cite{thompson-post-2020-automatic, thompson-post-2020-paraphrase} \\
chrF++ & \cite{popovic-2017-chrf} & Random-sysname & \cite{freitag-etal-2023-results} \\
chrfS & \cite{mukherjee-shrivastava-2024-chrf} & REUSE & \cite{mukherjee-shrivastava-2022-reuse} \\
COMET & \cite{rei-etal-2020-unbabels} & sentBLEU & \cite{papineni-etal-2002-bleu} \\
COMET-20 & \cite{rei-etal-2020-comet} & sentinel-cand-mqm & \cite{perrella-etal-2024-guardians} \\
COMET-22 & \cite{rei-etal-2022-comet} & sentinel-ref-mqm & \cite{perrella-etal-2024-guardians} \\
COMET-2R & \cite{rei-etal-2020-unbabels} & sentinel-src-mqm & \cite{perrella-etal-2024-guardians} \\
COMET-HTER & \cite{rei-etal-2020-unbabels} & SEScore & \cite{xu-etal-2022-errors} \\
COMET-MQM & \cite{rei-etal-2020-unbabels} & sescoreX & \cite{xu-etal-2023-sescore2} \\
COMET-QE & \cite{rei-etal-2021-references} & spBLEU & \cite{nllbteam2022languageleftbehindscaling} \\
COMET-Rank & \cite{rei-etal-2020-unbabels} & SWSS+METEOR & \cite{xu-etal-2020-incorporate} \\
COMETKiwi & \cite{rei-etal-2022-cometkiwi} & TER & \cite{snover-etal-2006-study} \\
CometKiwi-XL & \cite{rei-etal-2023-scaling} & tokengram\_F & \cite{dreano-etal-2023-tokengram} \\
CometKiwi-XXL & \cite{rei-etal-2023-scaling} & UniTE & \cite{wan-etal-2022-unite, wan-etal-2022-alibaba} \\
cometoid22-wmt22 & \cite{gowda-etal-2023-cometoid} & UniTE-src & \cite{wan-etal-2022-unite} \\
damonmonli & \cite{freitag-etal-2024-llms} & XCOMET & \cite{guerreiro-etal-2024-xcomet} \\
docWMT22CometDA & \cite{vernikos-etal-2022-embarrassingly} & XCOMET-Ensemble & \cite{guerreiro-etal-2024-xcomet} \\
docWMT22CometKiwiDA & \cite{vernikos-etal-2022-embarrassingly} & XCOMET-QE & \cite{guerreiro-etal-2024-xcomet} \\
eBLEU & \cite{elnokrashy-kocmi-2023-ebleu} & XCOMET-QE-Ensemble & \cite{guerreiro-etal-2024-xcomet} \\
EED & \cite{stanchev-etal-2019-eed} & XLsim & \cite{mukherjee-shrivastava-2023-mee4} \\
embed\_llama & \cite{dreano-etal-2023-embed} & XLsimMqm & \cite{mukherjee-shrivastava-2023-mee4} \\
esim & \cite{mathur-etal-2019-putting} & YiSi-0 & \cite{lo-2019-yisi} \\
f200spBLEU & \cite{nllbteam2022languageleftbehindscaling} & YiSi-1 & \cite{lo-2019-yisi} \\
GEMBA-MQM & \cite{kocmi-federmann-2023-gemba} & YiSi-2 & \cite{lo-2019-yisi} \\
gemba\_esa & \cite{freitag-etal-2024-llms} & Yisi-combi & \cite{mathur-etal-2020-results} \\
HWTSC-Teacher-Sim & \cite{liu-etal-2022-partial} & yisi1-translate & \cite{mathur-etal-2020-results} \\
KG-BERTScore & \cite{wu-etal-2023-empowering} & mbr-metricx-qe & \cite{naskar-etal-2023-quality} \\
MaTESe & \cite{perrella-etal-2022-matese} & mBERT-L2 & \cite{mathur-etal-2020-results} \\
MaTESe-QE & \cite{perrella-etal-2022-matese} & MEE & \cite{mee-metric} \\
MEE4 & \cite{mukherjee-shrivastava-2022-unsupervised} & & \\
\bottomrule
\end{tabular}
}
\caption{List of all automatic evaluators considered, i.e., MT metrics, associated with their reference papers. Metrics without dedicated papers cite the Metrics Shared Task results paper in which they appeared.}
\label{tab:metrics}
\end{table*}

%% file: tables/wmt2020.tex
\begin{table*}[t]
\centering
\begin{tabular}{lrr|rr|rr|rr}
\toprule
& \multicolumn{4}{c|}{\langpair{en}{de}}& \multicolumn{4}{c}{\langpair{zh}{en}}\\
& \multicolumn{2}{c|}{SPA}& \multicolumn{2}{c|}{\acceq}& \multicolumn{2}{c|}{SPA}& \multicolumn{2}{c|}{\acceq}\\
\multicolumn{1}{l}{Metric} & \multicolumn{1}{c}{ Rank } & \multicolumn{1}{c|}{ Acc. } & \multicolumn{1}{c}{ Rank } & \multicolumn{1}{c|}{ Acc. } & \multicolumn{1}{c}{ Rank } & \multicolumn{1}{c|}{ Acc. } & \multicolumn{1}{c}{ Rank } & \multicolumn{1}{c|}{ Acc. }\\
\midrule
\cellcolor{highlight}MQM-2020-2& \cellcolor{highlight} $1$ & \cellcolor{highlight} $96.45$& \cellcolor{highlight} $1$ & \cellcolor{highlight} $58.86$& \cellcolor{highlight} $1$ & \cellcolor{highlight} $88.10$& \cellcolor{highlight} $1$ & \cellcolor{highlight} $55.70$\\
\cellcolor{highlight}pSQM-1& \cellcolor{highlight} $1$ & \cellcolor{highlight} $95.59$& \cellcolor{highlight} $6$ & \cellcolor{highlight} $49.41$& \cellcolor{highlight} $1$ & \cellcolor{highlight} $79.16$& \cellcolor{highlight} $13$ & \cellcolor{highlight} $43.89$\\
\cellcolor{highlight}MQM-2020-3& \cellcolor{highlight} $2$ & \cellcolor{highlight} $90.39$& \cellcolor{highlight} $2$ & \cellcolor{highlight} $56.84$& \cellcolor{highlight} $1$ & \cellcolor{highlight} $92.06$& \cellcolor{highlight} $2$ & \cellcolor{highlight} $52.80$\\
BLEURT-0.2& $2$ & $86.81$& $4$ & $50.81$& $2$ & $72.59$& $3$ & $50.57$\\
\cellcolor{highlight}pSQM-2& \cellcolor{highlight} $2$ & \cellcolor{highlight} $85.87$& \cellcolor{highlight} $9$ & \cellcolor{highlight} $46.97$& \cellcolor{highlight} $1$ & \cellcolor{highlight} $89.33$& \cellcolor{highlight} $9$ & \cellcolor{highlight} $46.77$\\
BLEURT-20& $2$ & $85.52$& $3$ & $51.68$& $3$ & $67.46$& $4$ & $50.12$\\
\cellcolor{highlight}pSQM-3& \cellcolor{highlight} $2$ & \cellcolor{highlight} $84.61$& \cellcolor{highlight} $6$ & \cellcolor{highlight} $49.38$& \cellcolor{highlight} $1$ & \cellcolor{highlight} $87.94$& \cellcolor{highlight} $7$ & \cellcolor{highlight} $47.88$\\
all-rembert-20& $3$ & $79.19$& $4$ & $51.04$& $3$ & $66.41$& $3$ & $50.61$\\
BLEURT-extended& $3$ & $75.55$& $5$ & $50.21$& $3$ & $64.00$& $3$ & $50.74$\\
COMET-MQM& $4$ & $71.39$& $7$ & $48.21$& $4$ & $55.43$& $6$ & $48.49$\\
BLEURT-0.1-all& $4$ & $71.38$& $7$ & $48.63$& $2$ & $71.04$& $5$ & $49.54$\\
COMET& $4$ & $71.09$& $8$ & $47.36$& $4$ & $56.01$& $5$ & $49.28$\\
COMET-QE*& $4$ & $70.59$& $8$ & $47.82$& $3$ & $58.37$& $8$ & $47.09$\\
COMET-HTER& $5$ & $65.71$& $8$ & $47.62$& $4$ & $54.79$& $5$ & $49.30$\\
mBERT-L2& $5$ & $65.03$& $10$ & $45.48$& $4$ & $56.49$& $6$ & $48.97$\\
COMET-2R& $6$ & $58.12$& $9$ & $46.43$& $4$ & $55.99$& $4$ & $50.20$\\
COMET-Rank& $6$ & $54.78$& $14$ & $41.31$& $3$ & $58.16$& $14$ & $43.57$\\
OpenKiwi-XLMR*& $6$ & $53.25$& $11$ & $44.11$& $4$ & $53.29$& $8$ & $47.23$\\
OpenKiwi-Bert*& $6$ & $52.01$& $16$ & $39.98$& $3$ & $59.55$& $11$ & $45.13$\\
prism& $6$ & $51.92$& $11$ & $43.59$& $4$ & $57.88$& $8$ & $47.56$\\
Yisi-combi& $7$ & $51.10$& $12$ & $42.63$& -- & --& -- & --\\
bleurt-combi& $7$ & $51.10$& $12$ & $42.63$& -- & --& -- & --\\
esim& $7$ & $50.72$& $14$ & $41.35$& $4$ & $52.90$& $10$ & $46.19$\\
chrF& $7$ & $49.86$& $13$ & $42.05$& $5$ & $47.70$& $13$ & $44.09$\\
EED& $7$ & $49.81$& $15$ & $40.94$& $5$ & $45.41$& $14$ & $43.64$\\
paresim-1& $7$ & $49.54$& $14$ & $41.37$& $4$ & $53.34$& $10$ & $46.15$\\
chrF++& $7$ & $48.87$& $13$ & $41.99$& $5$ & $48.96$& $12$ & $44.27$\\
YiSi-1& $7$ & $48.79$& $12$ & $42.70$& $4$ & $52.74$& $7$ & $48.01$\\
CharacTER& $7$ & $47.71$& $16$ & $40.45$& $5$ & $48.84$& $13$ & $44.01$\\
BLEURT-0.1-en& $7$ & $47.43$& $15$ & $40.96$& $4$ & $57.29$& $7$ & $48.26$\\
YiSi-0& $7$ & $46.23$& $17$ & $39.78$& $5$ & $46.47$& $14$ & $43.60$\\
TER& $7$ & $45.98$& $16$ & $40.15$& $6$ & $39.68$& $15$ & $43.34$\\
parchrf++& $7$ & $45.57$& $13$ & $42.25$& $5$ & $48.68$& $12$ & $44.25$\\
MEE& $7$ & $45.31$& $14$ & $41.61$& $4$ & $52.91$& $13$ & $43.94$\\
sentBLEU& $7$ & $44.41$& $15$ & $41.07$& $4$ & $50.45$& $15$ & $43.37$\\
parbleu& $8$ & $41.38$& $15$ & $41.01$& $4$ & $50.28$& $15$ & $43.43$\\
yisi1-translate& $8$ & $39.76$& $12$ & $42.60$& $4$ & $52.28$& $11$ & $44.70$\\
YiSi-2*& $8$ & $38.44$& $18$ & $34.36$& $5$ & $43.35$& $12$ & $44.60$\\
\bottomrule
\end{tabular}

\caption{2020}
\label{tab:full-wmt2020}
\end{table*}

%% file: tables/wmt2022.tex
\begin{table*}[t]
    \centering
\begin{tabular}{lrr|rr|rr|rr}
\toprule
& \multicolumn{4}{c|}{\langpair{en}{de}}& \multicolumn{4}{c}{\langpair{en}{zh}}\\
& \multicolumn{2}{c|}{SPA}& \multicolumn{2}{c|}{\acceq}& \multicolumn{2}{c|}{SPA}& \multicolumn{2}{c|}{\acceq}\\
\multicolumn{1}{l}{Metric} & \multicolumn{1}{c}{ Rank } & \multicolumn{1}{c|}{ Acc. } & \multicolumn{1}{c}{ Rank } & \multicolumn{1}{c|}{ Acc. } & \multicolumn{1}{c}{ Rank } & \multicolumn{1}{c|}{ Acc. } & \multicolumn{1}{c}{ Rank } & \multicolumn{1}{c|}{ Acc. }\\
\midrule
MetricX-23-QE-XXL*& $1$ & $94.89$& $3$ & $57.64$& $2$ & $83.92$& $2$ & $47.43$\\
\cellcolor{highlight}MQM-2022-2& \cellcolor{highlight} $1$ & \cellcolor{highlight} $94.49$& \cellcolor{highlight} $6$ & \cellcolor{highlight} $55.55$& \cellcolor{highlight} $2$ & \cellcolor{highlight} $80.82$& \cellcolor{highlight} $3$ & \cellcolor{highlight} $47.05$\\
\cellcolor{highlight}MQM-2022-3& \cellcolor{highlight} $1$ & \cellcolor{highlight} $92.59$& \cellcolor{highlight} $1$ & \cellcolor{highlight} $61.06$& \cellcolor{highlight} $1$ & \cellcolor{highlight} $87.22$& \cellcolor{highlight} $2$ & \cellcolor{highlight} $47.56$\\
MetricX-23-XXL& $2$ & $92.34$& $2$ & $59.27$& $1$ & $87.69$& $1$ & $48.43$\\
COMET-22& $2$ & $91.63$& $5$ & $56.51$& $2$ & $84.08$& $3$ & $46.74$\\
COMET-20& $2$ & $91.28$& $9$ & $52.42$& $2$ & $80.56$& $7$ & $43.81$\\
CometKiwi*& $2$ & $89.51$& $7$ & $53.77$& $3$ & $75.36$& $8$ & $43.21$\\
BLEURT-20& $3$ & $88.20$& $7$ & $53.33$& $3$ & $77.80$& $7$ & $43.84$\\
metricx\_xxl\_MQM\_2020& $3$ & $88.10$& $3$ & $57.43$& $1$ & $87.04$& $3$ & $46.89$\\
COMET-QE*& $3$ & $85.51$& $10$ & $51.69$& $3$ & $78.33$& $7$ & $43.61$\\
MS-COMET-22& $3$ & $85.37$& $8$ & $53.13$& $1$ & $85.18$& $6$ & $44.92$\\
CometKiwi-XXL*& $3$ & $84.43$& $7$ & $53.27$& $2$ & $81.25$& $2$ & $47.28$\\
UniTE& $4$ & $82.77$& $4$ & $57.03$& $2$ & $83.88$& $5$ & $45.86$\\
UniTE-src*& $4$ & $81.55$& $6$ & $55.00$& $4$ & $65.74$& $7$ & $43.53$\\
CometKiwi-XL*& $4$ & $81.13$& $8$ & $52.73$& $2$ & $81.56$& $4$ & $46.33$\\
YiSi-1& $4$ & $78.91$& $13$ & $48.26$& $4$ & $70.72$& $8$ & $43.23$\\
MATESE& $5$ & $78.03$& $7$ & $53.48$& -- & --& -- & --\\
BERTScore& $5$ & $75.61$& $14$ & $47.57$& $4$ & $70.69$& $8$ & $43.28$\\
SEScore& $5$ & $75.16$& $12$ & $50.45$& -- & --& -- & --\\
MS-COMET-QE-22*& $5$ & $74.44$& $12$ & $50.37$& $2$ & $78.84$& $9$ & $42.51$\\
MEE4& $5$ & $74.19$& $15$ & $46.81$& -- & --& -- & --\\
chrF& $5$ & $73.05$& $16$ & $46.38$& $3$ & $72.67$& $10$ & $41.87$\\
f200spBLEU& $5$ & $71.04$& $15$ & $46.84$& $4$ & $71.76$& $10$ & $41.85$\\
HWTSC-Teacher-Sim*& $5$ & $69.68$& $13$ & $48.10$& $4$ & $68.43$& $11$ & $40.53$\\
\cellcolor{highlight}DA+SQM& \cellcolor{highlight} $6$ & \cellcolor{highlight} $66.61$& \cellcolor{highlight} $16$ & \cellcolor{highlight} $46.03$& \cellcolor{highlight} $2$ & \cellcolor{highlight} $82.95$& \cellcolor{highlight} $12$ & \cellcolor{highlight} $36.26$\\
MATESE-QE*& $6$ & $65.42$& $11$ & $51.06$& -- & --& -- & --\\
BLEU& $6$ & $65.00$& $15$ & $46.51$& $4$ & $67.31$& $13$ & $34.28$\\
REUSE*& $7$ & $37.95$& $17$ & $43.58$& $5$ & $33.46$& $12$ & $35.89$\\
\bottomrule
\end{tabular}
\caption{2022}
\label{tab:full-wmt2022}
\end{table*}

%% file: tables/wmt2023.tex
\begin{table*}[t]
\centering

\begin{tabular}{lrr|rr|rr|rr}
\toprule
& \multicolumn{4}{c|}{\langpair{en}{de}}& \multicolumn{4}{c}{\langpair{zh}{en}}\\
& \multicolumn{2}{c|}{SPA}& \multicolumn{2}{c|}{\acceq}& \multicolumn{2}{c|}{SPA}& \multicolumn{2}{c|}{\acceq}\\
\multicolumn{1}{l}{Metric} & \multicolumn{1}{c}{ Rank } & \multicolumn{1}{c|}{ Acc. } & \multicolumn{1}{c}{ Rank } & \multicolumn{1}{c|}{ Acc. } & \multicolumn{1}{c}{ Rank } & \multicolumn{1}{c|}{ Acc. } & \multicolumn{1}{c}{ Rank } & \multicolumn{1}{c|}{ Acc. }\\
\midrule
GEMBA-MQM*& $1$ & $94.52$& $5$ & $58.52$& $1$ & $93.17$& $3$ & $52.80$\\
\cellcolor{highlight}MQM-2023-3& \cellcolor{highlight} $1$ & \cellcolor{highlight} $93.51$& \cellcolor{highlight} $5$ & \cellcolor{highlight} $58.42$& \cellcolor{highlight} $1$ & \cellcolor{highlight} $95.54$& \cellcolor{highlight} $5$ & \cellcolor{highlight} $51.65$\\
CometKiwi-XXL*& $1$ & $93.22$& $5$ & $58.46$& $1$ & $92.86$& $6$ & $50.94$\\
\cellcolor{highlight}MQM-2023-2& \cellcolor{highlight} $1$ & \cellcolor{highlight} $93.15$& \cellcolor{highlight} $6$ & \cellcolor{highlight} $57.71$& \cellcolor{highlight} $1$ & \cellcolor{highlight} $95.18$& \cellcolor{highlight} $2$ & \cellcolor{highlight} $52.90$\\
CometKiwi-XL*& $1$ & $93.11$& $6$ & $57.38$& $2$ & $92.02$& $6$ & $50.71$\\
MetricX-23-XXL& $1$ & $92.57$& $2$ & $61.82$& $2$ & $91.58$& $2$ & $53.13$\\
XCOMET-QE-Ensemble*& $1$ & $92.48$& $4$ & $59.89$& $2$ & $90.54$& $3$ & $52.87$\\
cometoid22-wmt22*& $1$ & $92.43$& $5$ & $58.09$& $2$ & $90.09$& $7$ & $50.23$\\
COMET& $1$ & $92.33$& $7$ & $56.65$& $4$ & $87.18$& $9$ & $48.42$\\
XCOMET-Ensemble& $1$ & $92.21$& $3$ & $60.99$& $2$ & $91.15$& $1$ & $54.59$\\
MetricX-23-QE-XXL*& $1$ & $92.12$& $1$ & $62.53$& $3$ & $88.30$& $2$ & $53.26$\\
Calibri-COMET22& $1$ & $92.01$& $10$ & $51.26$& $4$ & $87.02$& $16$ & $44.57$\\
docWMT22CometDA& $1$ & $91.76$& $8$ & $54.71$& $4$ & $87.41$& $13$ & $46.15$\\
sescoreX& $1$ & $91.66$& $8$ & $54.76$& $4$ & $85.73$& $13$ & $46.39$\\
\cellcolor{highlight}DA+SQM& \cellcolor{highlight} $2$ & \cellcolor{highlight} $91.24$& \cellcolor{highlight} $14$ & \cellcolor{highlight} $46.79$& \cellcolor{highlight} $4$ & \cellcolor{highlight} $86.28$& \cellcolor{highlight} $22$ & \cellcolor{highlight} $39.42$\\
Calibri-COMET22-QE*& $2$ & $90.80$& $11$ & $50.33$& $4$ & $87.59$& $11$ & $47.24$\\
\cellcolor{highlight}ESA-1& \cellcolor{highlight} $2$ & \cellcolor{highlight} $90.39$& \cellcolor{highlight} $14$ & \cellcolor{highlight} $46.71$& \cellcolor{highlight} -- & \cellcolor{highlight} --& \cellcolor{highlight} -- & \cellcolor{highlight} --\\
BLEURT-20& $2$ & $90.35$& $8$ & $55.19$& $4$ & $87.36$& $9$ & $48.63$\\
mbr-metricx-qe*& $2$ & $89.98$& $5$ & $58.75$& $3$ & $88.55$& $4$ & $52.04$\\
prismRef& $2$ & $89.92$& $11$ & $50.72$& $5$ & $82.50$& $14$ & $46.06$\\
docWMT22CometKiwiDA*& $2$ & $89.92$& $8$ & $55.30$& $2$ & $90.95$& $10$ & $47.83$\\
MS-COMET-QE-22*& $2$ & $89.85$& $8$ & $54.55$& $4$ & $87.59$& $10$ & $47.81$\\
f200spBLEU& $2$ & $89.24$& $11$ & $50.54$& $5$ & $81.12$& $18$ & $43.33$\\
CometKiwi*& $2$ & $89.23$& $6$ & $57.74$& $3$ & $89.37$& $5$ & $51.74$\\
mre-score-labse-regular& $2$ & $89.14$& $10$ & $51.12$& $4$ & $87.14$& $17$ & $43.80$\\
\cellcolor{highlight}ESA-2& \cellcolor{highlight} $2$ & \cellcolor{highlight} $89.11$& \cellcolor{highlight} $12$ & \cellcolor{highlight} $49.70$& \cellcolor{highlight} -- & \cellcolor{highlight} --& \cellcolor{highlight} -- & \cellcolor{highlight} --\\
YiSi-1& $2$ & $88.96$& $9$ & $53.15$& $4$ & $85.70$& $12$ & $46.68$\\
\cellcolor{highlight}MQM-2023-4& \cellcolor{highlight} $2$ & \cellcolor{highlight} $88.93$& \cellcolor{highlight} $14$ & \cellcolor{highlight} $46.68$& \cellcolor{highlight} -- & \cellcolor{highlight} --& \cellcolor{highlight} -- & \cellcolor{highlight} --\\
KG-BERTScore*& $2$ & $88.79$& $7$ & $56.98$& $3$ & $89.31$& $8$ & $49.75$\\
MaTESe& $2$ & $88.40$& $9$ & $53.36$& $2$ & $92.06$& $7$ & $50.34$\\
BLEU& $2$ & $88.02$& $12$ & $50.06$& $6$ & $80.92$& $19$ & $43.13$\\
BERTscore& $2$ & $87.33$& $11$ & $50.88$& $5$ & $84.68$& $15$ & $45.79$\\
MEE4& $2$ & $87.07$& $10$ & $51.62$& $6$ & $80.51$& $19$ & $42.94$\\
XLsim& $2$ & $86.58$& $10$ & $51.01$& $6$ & $81.00$& $19$ & $42.84$\\
tokengram\_F& $3$ & $85.60$& $12$ & $49.72$& $5$ & $81.01$& $18$ & $43.52$\\
chrF& $4$ & $84.25$& $12$ & $49.54$& $5$ & $81.47$& $17$ & $43.72$\\
eBLEU& $4$ & $83.87$& $13$ & $48.96$& $6$ & $80.44$& $20$ & $42.55$\\
embed\_llama& $4$ & $81.33$& $14$ & $47.12$& $4$ & $84.84$& $21$ & $41.05$\\
Random-sysname*& $5$ & $59.47$& $16$ & $39.07$& $7$ & $54.34$& $23$ & $34.49$\\
prismSrc*& $6$ & $30.03$& $15$ & $40.89$& $8$ & $35.54$& $22$ & $39.28$\\
\bottomrule
\end{tabular}

\caption{2023}
\label{tab:full-wmt2023}
\end{table*}

%% file: tables/wmt2024.tex
\begin{table*}[t]
    \centering
    \begin{tabular}{lrr|rr}
    \toprule
    & \multicolumn{4}{c}{\langpair{en}{es}}\\
    & \multicolumn{2}{c|}{SPA}& \multicolumn{2}{c}{\acceq}\\
    \multicolumn{1}{l}{Metric} & \multicolumn{1}{c}{ Rank } & \multicolumn{1}{c|}{ Acc. } & \multicolumn{1}{c}{ Rank } & \multicolumn{1}{c}{ Acc. }\\
    \midrule
    CometKiwi-XXL*& $1$ & $86.12$& $4$ & $67.24$\\
    gemba\_esa*& $1$ & $85.72$& $3$ & $67.68$\\
    COMET-22& $1$ & $82.37$& $5$ & $66.60$\\
    bright-qe*& $1$ & $81.77$& $4$ & $67.39$\\
    \cellcolor{highlight}ESA& \cellcolor{highlight} $2$ & \cellcolor{highlight} $80.12$& \cellcolor{highlight} $8$ & \cellcolor{highlight} $63.84$\\
    XCOMET-QE*& $2$ & $80.10$& $3$ & $67.99$\\
    metametrics\_mt\_mqm\_hybrid\_kendall& $2$ & $80.10$& $1$ & $68.95$\\
    XCOMET& $2$ & $79.96$& $2$ & $68.67$\\
    MetricX-24-Hybrid& $2$ & $79.75$& $1$ & $69.20$\\
    BLCOM\_1& $2$ & $79.17$& $6$ & $65.02$\\
    MetricX-24-Hybrid-QE*& $2$ & $79.09$& $2$ & $68.92$\\
    sentinel-cand-mqm*& $2$ & $78.54$& $5$ & $66.39$\\
    BLEURT-20& $2$ & $75.96$& $7$ & $64.48$\\
    metametrics\_mt\_mqm\_qe\_kendall.seg.s*& $3$ & $73.29$& $4$ & $67.49$\\
    CometKiwi*& $3$ & $71.74$& $5$ & $66.51$\\
    PrismRefMedium& $3$ & $70.93$& $11$ & $61.39$\\
    PrismRefSmall& $3$ & $70.52$& $10$ & $61.51$\\
    YiSi-1& $3$ & $70.51$& $11$ & $61.44$\\
    BERTScore& $3$ & $67.75$& $11$ & $61.41$\\
    chrF& $3$ & $66.73$& $13$ & $61.05$\\
    damonmonli& $3$ & $66.37$& $9$ & $62.10$\\
    chrfS& $4$ & $64.31$& $11$ & $61.37$\\
    spBLEU& $4$ & $63.19$& $12$ & $61.08$\\
    BLEU& $5$ & $60.67$& $13$ & $61.04$\\
    MEE4& $5$ & $60.36$& $10$ & $61.57$\\
    sentinel-ref-mqm& $6$ & $44.19$& $13$ & $61.04$\\
    sentinel-src-mqm*& $6$ & $44.19$& $13$ & $61.04$\\
    XLsimMqm*& $6$ & $39.25$& $12$ & $61.11$\\
    \bottomrule
    \end{tabular}
    
    \caption{2024}
    \label{tab:full-wmt24}
\end{table*}

%% file: tables/varying_ground_truth/wmt2020_gold_psqm.tex
\begin{table*}[t]
    \centering

    \begin{tabular}{lrr|rr}
\toprule
& \multicolumn{4}{c}{\langpair{zh}{en}}\\
& \multicolumn{2}{c|}{SPA}& \multicolumn{2}{c}{\acceq}\\
\multicolumn{1}{l}{Metric} & \multicolumn{1}{c}{ Rank } & \multicolumn{1}{c|}{ Acc. } & \multicolumn{1}{c}{ Rank } & \multicolumn{1}{c}{ Acc. }\\
\midrule
\cellcolor{highlight}pSQM-3& \cellcolor{highlight} $1$ & \cellcolor{highlight} $83.42$& \cellcolor{highlight} $6$ & \cellcolor{highlight} $65.23$\\
\cellcolor{highlight}MQM-2020-2& \cellcolor{highlight} $1$ & \cellcolor{highlight} $80.56$& \cellcolor{highlight} $6$ & \cellcolor{highlight} $65.22$\\
\cellcolor{highlight}MQM-2020-3& \cellcolor{highlight} $1$ & \cellcolor{highlight} $80.34$& \cellcolor{highlight} $6$ & \cellcolor{highlight} $65.22$\\
\cellcolor{highlight}MQM-2020-1& \cellcolor{highlight} $1$ & \cellcolor{highlight} $79.55$& \cellcolor{highlight} $6$ & \cellcolor{highlight} $65.22$\\
\cellcolor{highlight}pSQM-2& \cellcolor{highlight} $1$ & \cellcolor{highlight} $74.02$& \cellcolor{highlight} $4$ & \cellcolor{highlight} $65.26$\\
BERT-large-L2& $1$ & $70.19$& $3$ & $65.38$\\
COMET& $1$ & $68.58$& $1$ & $65.64$\\
SWSS+METEOR& $2$ & $67.26$& $4$ & $65.26$\\
MEE& $2$ & $67.07$& $6$ & $65.22$\\
prism& $2$ & $66.01$& $5$ & $65.25$\\
sentBLEU& $2$ & $65.70$& $5$ & $65.25$\\
parbleu& $2$ & $65.63$& $6$ & $65.23$\\
BLEURT& $2$ & $65.28$& $2$ & $65.49$\\
YiSi-1& $2$ & $64.71$& $6$ & $65.22$\\
yisi1-translate& $2$ & $64.32$& $6$ & $65.23$\\
CharacTER& $2$ & $63.99$& $6$ & $65.23$\\
all-rembert-20& $2$ & $63.48$& $2$ & $65.47$\\
BLEURT-20& $2$ & $63.10$& $3$ & $65.38$\\
paresim-1& $2$ & $62.79$& $4$ & $65.30$\\
esim& $2$ & $62.42$& $4$ & $65.30$\\
chrF++& $3$ & $62.38$& $6$ & $65.23$\\
BLEURT-0.1-en& $3$ & $62.22$& $2$ & $65.47$\\
COMET-2R& $3$ & $62.10$& $1$ & $65.61$\\
parchrf++& $3$ & $61.92$& $5$ & $65.24$\\
EED& $3$ & $60.91$& $6$ & $65.23$\\
mBERT-L2& $3$ & $60.84$& $2$ & $65.43$\\
chrF& $3$ & $60.81$& $4$ & $65.25$\\
YiSi-0& $3$ & $60.55$& $5$ & $65.25$\\
BLEURT-0.2& $3$ & $60.54$& $3$ & $65.34$\\
COMET-Rank& $3$ & $59.97$& $6$ & $65.22$\\
BLEURT-extended& $3$ & $59.90$& $3$ & $65.36$\\
BAQ\_static& $3$ & $59.89$& $6$ & $65.22$\\
BLEURT-0.1-all& $3$ & $59.80$& $4$ & $65.25$\\
BERT-base-L2& $3$ & $59.57$& $2$ & $65.49$\\
COMET-HTER& $3$ & $58.88$& $2$ & $65.42$\\
COMET-MQM& $3$ & $58.27$& $5$ & $65.25$\\
BAQ\_dyn& $3$ & $58.12$& $6$ & $65.22$\\
COMET-QE*& $3$ & $57.83$& $4$ & $65.26$\\
TER& $3$ & $56.12$& $5$ & $65.25$\\
OpenKiwi-Bert*& $3$ & $55.35$& $5$ & $65.25$\\
OpenKiwi-XLMR*& $3$ & $50.98$& $4$ & $65.30$\\
YiSi-2*& $4$ & $48.35$& $4$ & $65.27$\\
\bottomrule
\end{tabular}

    \caption{The test set is 2020 \langpair{zh}{en}. The evaluator selected as the ground truth follows the pSQM protocol (pSQM-1 in Table~\ref{tab:full-wmt2020}).}
    \label{tab:wmt20-gold-psqm}
\end{table*}

%% file: tables/varying_ground_truth/wmt2023_gold_dasqm.tex
\begin{table*}[t]
    \centering

\begin{tabular}{lrr|rr}
\toprule
& \multicolumn{4}{c}{\langpair{en}{de}}\\
& \multicolumn{2}{c|}{SPA}& \multicolumn{2}{c}{\acceq}\\
\multicolumn{1}{l}{Metric} & \multicolumn{1}{c}{ Rank } & \multicolumn{1}{c|}{ Acc. } & \multicolumn{1}{c}{ Rank } & \multicolumn{1}{c}{ Acc. }\\
\midrule
MetricX-23-QE-XXL*& $1$ & $95.11$& $3$ & $58.41$\\
GEMBA-MQM*& $1$ & $94.89$& $14$ & $43.06$\\
CometKiwi*& $1$ & $94.83$& $3$ & $58.21$\\
KG-BERTScore*& $1$ & $94.57$& $5$ & $57.16$\\
\cellcolor{highlight}MQM-2023-3& \cellcolor{highlight} $1$ & \cellcolor{highlight} $94.24$& \cellcolor{highlight} $13$ & \cellcolor{highlight} $46.99$\\
docWMT22CometKiwiDA*& $1$ & $94.22$& $2$ & $58.60$\\
CometKiwi-XL*& $1$ & $94.16$& $2$ & $58.80$\\
MS-COMET-QE-22*& $1$ & $94.04$& $7$ & $55.66$\\
CometKiwi-XXL*& $1$ & $93.98$& $1$ & $59.66$\\
MetricX-23-XXL& $2$ & $93.31$& $3$ & $58.06$\\
COMET& $2$ & $92.80$& $3$ & $58.42$\\
docWMT22CometDA& $2$ & $92.45$& $2$ & $58.95$\\
mre-score-labse-regular& $2$ & $92.30$& $8$ & $54.75$\\
\cellcolor{highlight}MQM-2023-1& \cellcolor{highlight} $2$ & \cellcolor{highlight} $92.10$& \cellcolor{highlight} $15$ & \cellcolor{highlight} $42.20$\\
Calibri-COMET22& $2$ & $92.02$& $3$ & $58.17$\\
mbr-metricx-qe*& $2$ & $91.94$& $3$ & $58.26$\\
\cellcolor{highlight}MQM-2023-2& \cellcolor{highlight} $2$ & \cellcolor{highlight} $91.40$& \cellcolor{highlight} $11$ & \cellcolor{highlight} $48.91$\\
cometoid22-wmt22*& $2$ & $91.34$& $5$ & $57.06$\\
sescoreX& $2$ & $91.22$& $4$ & $57.41$\\
BLEURT-20& $3$ & $90.66$& $5$ & $57.26$\\
prismRef& $3$ & $89.58$& $10$ & $54.16$\\
Calibri-COMET22-QE*& $3$ & $89.55$& $8$ & $55.22$\\
YiSi-1& $3$ & $89.41$& $6$ & $56.64$\\
XLsim& $3$ & $87.93$& $6$ & $56.47$\\
XCOMET-Ensemble& $4$ & $87.77$& $4$ & $57.82$\\
XCOMET-QE-Ensemble*& $4$ & $87.34$& $6$ & $56.40$\\
eBLEU& $4$ & $87.33$& $10$ & $53.91$\\
BERTscore& $4$ & $86.83$& $7$ & $56.13$\\
f200spBLEU& $4$ & $86.82$& $8$ & $54.97$\\
MaTESe& $4$ & $86.69$& $16$ & $37.35$\\
MEE4& $4$ & $86.12$& $7$ & $55.93$\\
BLEU& $5$ & $84.45$& $10$ & $53.63$\\
tokengram\_F& $5$ & $83.15$& $8$ & $54.98$\\
chrF& $5$ & $82.45$& $9$ & $54.74$\\
embed\_llama& $5$ & $81.27$& $9$ & $54.28$\\
Random-sysname*& $6$ & $60.55$& $12$ & $47.94$\\
prismSrc*& $7$ & $28.59$& $11$ & $48.54$\\
\bottomrule
\end{tabular}

    \caption{The test set is 2023 \langpair{en}{de}. The evaluator selected as the ground truth follows the DA+SQM protocol (DA+SQM in Table~\ref{tab:full-wmt2023}). Different from Tables~\ref{tab:rankings} and \ref{tab:full-wmt2023}, we exclude the evaluators ESA-1, ESA-2, and MQM-2023-4, because they annotated a limited number of translations. This way, we increase the number of segments in the test set from $145$ to $376$.}
    
    \label{tab:wmt23-gold-dasqm}
\end{table*}

%% file: tables/varying_ground_truth/wmt2023_gold_mqm1.tex
\begin{table*}[t]
    \centering

\begin{tabular}{lrr|rr}
\toprule
& \multicolumn{4}{c}{\langpair{en}{de}}\\
& \multicolumn{2}{c|}{SPA}& \multicolumn{2}{c}{\acceq}\\
\multicolumn{1}{l}{Metric} & \multicolumn{1}{c}{ Rank } & \multicolumn{1}{c|}{ Acc. } & \multicolumn{1}{c}{ Rank } & \multicolumn{1}{c}{ Acc. }\\
\midrule
\cellcolor{highlight}MQM-2023-3& \cellcolor{highlight} $1$ & \cellcolor{highlight} $97.09$& \cellcolor{highlight} $7$ & \cellcolor{highlight} $56.62$\\
GEMBA-MQM*& $1$ & $97.09$& $5$ & $59.18$\\
CometKiwi-XXL*& $1$ & $95.96$& $7$ & $56.43$\\
CometKiwi-XL*& $1$ & $95.47$& $6$ & $57.33$\\
\cellcolor{highlight}MQM-2023-2& \cellcolor{highlight} $1$ & \cellcolor{highlight} $95.20$& \cellcolor{highlight} $4$ & \cellcolor{highlight} $60.04$\\
docWMT22CometDA& $1$ & $94.99$& $10$ & $53.86$\\
MetricX-23-XXL& $2$ & $94.93$& $2$ & $61.65$\\
XCOMET-Ensemble& $2$ & $94.58$& $1$ & $62.23$\\
MetricX-23-QE-XXL*& $2$ & $94.52$& $3$ & $61.14$\\
COMET& $2$ & $94.39$& $8$ & $55.71$\\
XCOMET-QE-Ensemble*& $2$ & $94.14$& $4$ & $59.89$\\
docWMT22CometKiwiDA*& $2$ & $93.70$& $10$ & $53.55$\\
BLEURT-20& $2$ & $93.35$& $9$ & $54.83$\\
Calibri-COMET22-QE*& $2$ & $93.25$& $14$ & $49.62$\\
cometoid22-wmt22*& $2$ & $92.43$& $7$ & $56.73$\\
CometKiwi*& $3$ & $92.27$& $7$ & $56.76$\\
\cellcolor{highlight}DA+SQM& \cellcolor{highlight} $3$ & \cellcolor{highlight} $92.10$& \cellcolor{highlight} $18$ & \cellcolor{highlight} $43.68$\\
KG-BERTScore*& $3$ & $92.05$& $9$ & $54.40$\\
sescoreX& $3$ & $91.99$& $10$ & $53.99$\\
YiSi-1& $3$ & $91.28$& $11$ & $51.66$\\
mbr-metricx-qe*& $3$ & $91.22$& $8$ & $56.19$\\
MS-COMET-QE-22*& $3$ & $90.70$& $10$ & $53.51$\\
prismRef& $3$ & $90.22$& $14$ & $49.74$\\
Calibri-COMET22& $3$ & $88.28$& $14$ & $49.54$\\
XLsim& $4$ & $88.01$& $13$ & $50.31$\\
mre-score-labse-regular& $4$ & $87.47$& $13$ & $49.85$\\
BERTscore& $4$ & $87.01$& $12$ & $50.48$\\
f200spBLEU& $4$ & $86.90$& $13$ & $50.37$\\
MaTESe& $4$ & $86.73$& $7$ & $56.23$\\
eBLEU& $4$ & $86.24$& $16$ & $48.30$\\
MEE4& $4$ & $86.05$& $12$ & $50.81$\\
BLEU& $5$ & $84.48$& $15$ & $49.18$\\
tokengram\_F& $5$ & $83.85$& $14$ & $49.51$\\
chrF& $5$ & $83.34$& $14$ & $49.44$\\
embed\_llama& $5$ & $80.09$& $17$ & $45.05$\\
Random-sysname*& $6$ & $61.26$& $20$ & $38.19$\\
prismSrc*& $7$ & $28.88$& $19$ & $40.02$\\
\bottomrule
\end{tabular}

    \caption{The test set is 2023 \langpair{en}{de}. The evaluator selected as the ground truth follows the MQM protocol (it is the evaluator selected as ground truth in Table~\ref{tab:full-wmt2023}). Different from Tables~\ref{tab:rankings} and \ref{tab:full-wmt2023}, we exclude the evaluators ESA-1, ESA-2, and MQM-2023-4, because they annotated a limited number of translations. This way, we increase the number of segments in the test set from $145$ to $376$.}
    \label{tab:wmt23-gold-mqm1}
\end{table*}

%% file: tables/varying_ground_truth/wmt2023_gold_mqm2.tex
\begin{table*}[t]
    \centering

\begin{tabular}{lrr|rr}
\toprule
& \multicolumn{4}{c}{\langpair{en}{de}}\\
& \multicolumn{2}{c|}{SPA}& \multicolumn{2}{c}{\acceq}\\
\multicolumn{1}{l}{Metric} & \multicolumn{1}{c}{ Rank } & \multicolumn{1}{c|}{ Acc. } & \multicolumn{1}{c}{ Rank } & \multicolumn{1}{c}{ Acc. }\\
\midrule
BLEURT-20& $1$ & $97.12$& $6$ & $60.66$\\
MetricX-23-XXL& $1$ & $96.51$& $1$ & $64.61$\\
docWMT22CometDA& $1$ & $96.41$& $8$ & $59.16$\\
CometKiwi-XXL*& $1$ & $96.40$& $4$ & $61.57$\\
GEMBA-MQM*& $1$ & $96.19$& $8$ & $58.96$\\
CometKiwi-XL*& $1$ & $95.88$& $6$ & $60.45$\\
COMET& $1$ & $95.82$& $5$ & $61.31$\\
\cellcolor{highlight}MQM-2023-3& \cellcolor{highlight} $1$ & \cellcolor{highlight} $95.50$& \cellcolor{highlight} $8$ & \cellcolor{highlight} $59.25$\\
\cellcolor{highlight}MQM-2023-1& \cellcolor{highlight} $1$ & \cellcolor{highlight} $95.20$& \cellcolor{highlight} $7$ & \cellcolor{highlight} $60.03$\\
mbr-metricx-qe*& $1$ & $95.12$& $3$ & $62.39$\\
MetricX-23-QE-XXL*& $2$ & $94.99$& $2$ & $63.51$\\
Calibri-COMET22-QE*& $2$ & $94.48$& $15$ & $51.95$\\
docWMT22CometKiwiDA*& $2$ & $94.47$& $9$ & $57.58$\\
XCOMET-Ensemble& $2$ & $93.81$& $1$ & $64.20$\\
sescoreX& $2$ & $93.51$& $7$ & $59.94$\\
CometKiwi*& $2$ & $93.04$& $6$ & $60.24$\\
KG-BERTScore*& $2$ & $92.78$& $8$ & $59.06$\\
XCOMET-QE-Ensemble*& $3$ & $92.71$& $4$ & $61.92$\\
\cellcolor{highlight}DA+SQM& \cellcolor{highlight} $3$ & \cellcolor{highlight} $91.40$& \cellcolor{highlight} $16$ & \cellcolor{highlight} $48.92$\\
MS-COMET-QE-22*& $3$ & $91.34$& $9$ & $57.54$\\
cometoid22-wmt22*& $3$ & $91.22$& $5$ & $61.07$\\
YiSi-1& $3$ & $90.47$& $9$ & $57.50$\\
mre-score-labse-regular& $3$ & $90.02$& $10$ & $56.50$\\
prismRef& $3$ & $89.92$& $12$ & $55.16$\\
f200spBLEU& $3$ & $89.09$& $11$ & $55.81$\\
XLsim& $4$ & $88.77$& $11$ & $55.49$\\
Calibri-COMET22& $4$ & $88.37$& $11$ & $55.67$\\
eBLEU& $4$ & $88.19$& $14$ & $54.13$\\
BERTscore& $4$ & $88.09$& $10$ & $56.27$\\
BLEU& $4$ & $87.09$& $13$ & $54.61$\\
MaTESe& $4$ & $86.49$& $13$ & $54.19$\\
MEE4& $4$ & $86.46$& $10$ & $56.41$\\
tokengram\_F& $4$ & $86.12$& $11$ & $55.66$\\
chrF& $4$ & $85.53$& $11$ & $55.50$\\
embed\_llama& $5$ & $80.95$& $15$ & $51.39$\\
Random-sysname*& $6$ & $59.64$& $18$ & $42.43$\\
prismSrc*& $7$ & $25.26$& $17$ & $43.73$\\
\bottomrule
\end{tabular}

    \caption{The test set is 2023 \langpair{en}{de}. The evaluator selected as the ground truth follows the MQM protocol (MQM-2023-2 in Table~\ref{tab:full-wmt2023}). Different from Tables~\ref{tab:rankings} and \ref{tab:full-wmt2023}, we exclude the evaluators ESA-1, ESA-2, and MQM-2023-4, because they annotated a limited number of translations. This way, we increase the number of segments in the test set from $145$ to $376$.}
    \label{tab:wmt23-gold-mqm2}
\end{table*}

%% file: tables/varying_ground_truth/wmt2023_gold_mqm3.tex
\begin{table*}[t]
    \centering

\begin{tabular}{lrr|rr}
\toprule
& \multicolumn{4}{c}{\langpair{en}{de}}\\
& \multicolumn{2}{c|}{SPA}& \multicolumn{2}{c}{\acceq}\\
\multicolumn{1}{l}{Metric} & \multicolumn{1}{c}{ Rank } & \multicolumn{1}{c|}{ Acc. } & \multicolumn{1}{c}{ Rank } & \multicolumn{1}{c}{ Acc. }\\
\midrule
GEMBA-MQM*& $1$ & $97.98$& $7$ & $55.71$\\
CometKiwi-XXL*& $1$ & $97.96$& $4$ & $58.36$\\
CometKiwi-XL*& $1$ & $97.49$& $5$ & $57.78$\\
\cellcolor{highlight}MQM-2023-1& \cellcolor{highlight} $1$ & \cellcolor{highlight} $97.09$& \cellcolor{highlight} $6$ & \cellcolor{highlight} $56.46$\\
MetricX-23-XXL& $1$ & $96.86$& $1$ & $61.26$\\
docWMT22CometDA& $1$ & $96.82$& $6$ & $56.98$\\
MetricX-23-QE-XXL*& $1$ & $96.47$& $2$ & $60.55$\\
COMET& $2$ & $96.30$& $4$ & $58.63$\\
docWMT22CometKiwiDA*& $2$ & $96.09$& $8$ & $54.86$\\
\cellcolor{highlight}MQM-2023-2& \cellcolor{highlight} $2$ & \cellcolor{highlight} $95.50$& \cellcolor{highlight} $3$ & \cellcolor{highlight} $59.25$\\
CometKiwi*& $2$ & $94.67$& $5$ & $57.83$\\
KG-BERTScore*& $2$ & $94.45$& $6$ & $56.67$\\
sescoreX& $2$ & $94.31$& $5$ & $57.58$\\
\cellcolor{highlight}DA+SQM& \cellcolor{highlight} $2$ & \cellcolor{highlight} $94.24$& \cellcolor{highlight} $14$ & \cellcolor{highlight} $46.99$\\
BLEURT-20& $3$ & $94.18$& $4$ & $58.68$\\
Calibri-COMET22-QE*& $3$ & $94.01$& $13$ & $49.98$\\
YiSi-1& $3$ & $93.19$& $7$ & $55.64$\\
MS-COMET-QE-22*& $3$ & $93.17$& $7$ & $55.46$\\
mbr-metricx-qe*& $3$ & $93.06$& $3$ & $59.30$\\
XCOMET-Ensemble& $3$ & $92.64$& $2$ & $60.61$\\
prismRef& $3$ & $92.58$& $10$ & $53.57$\\
cometoid22-wmt22*& $3$ & $92.55$& $4$ & $58.50$\\
XCOMET-QE-Ensemble*& $3$ & $92.14$& $3$ & $59.43$\\
Calibri-COMET22& $4$ & $90.86$& $10$ & $53.18$\\
XLsim& $4$ & $90.86$& $8$ & $54.63$\\
BERTscore& $4$ & $89.67$& $9$ & $54.30$\\
f200spBLEU& $4$ & $89.56$& $9$ & $54.06$\\
mre-score-labse-regular& $4$ & $89.14$& $8$ & $54.47$\\
MEE4& $4$ & $88.72$& $8$ & $54.40$\\
eBLEU& $4$ & $88.67$& $11$ & $53.06$\\
BLEU& $5$ & $87.13$& $11$ & $52.76$\\
tokengram\_F& $5$ & $86.17$& $10$ & $53.62$\\
chrF& $5$ & $85.58$& $10$ & $53.51$\\
MaTESe& $5$ & $84.75$& $12$ & $52.12$\\
embed\_llama& $5$ & $82.93$& $13$ & $50.10$\\
Random-sysname*& $6$ & $58.48$& $16$ & $41.27$\\
prismSrc*& $7$ & $28.67$& $15$ & $44.52$\\
\bottomrule
\end{tabular}

    \caption{The test set is 2023 \langpair{en}{de}. The evaluator selected as the ground truth follows the MQM protocol (MQM-2023-3 in Table~\ref{tab:full-wmt2023}). Different from Tables~\ref{tab:rankings} and \ref{tab:full-wmt2023}, we exclude the evaluators ESA-1, ESA-2, and MQM-2023-4, because they annotated a limited number of translations. This way, we increase the number of segments in the test set from $145$ to $376$.}
    \label{tab:wmt23-gold-mqm3}
\end{table*}

%% file: tables/varying_ground_truth/wmt2024_gold_esa.tex
\begin{table*}[t]
    \centering
\begin{tabular}{lrr|rr}
\toprule
& \multicolumn{4}{c}{\langpair{en}{es}}\\
& \multicolumn{2}{c|}{SPA}& \multicolumn{2}{c}{\acceq}\\
\multicolumn{1}{l}{Metric} & \multicolumn{1}{c}{ Rank } & \multicolumn{1}{c|}{ Acc. } & \multicolumn{1}{c}{ Rank } & \multicolumn{1}{c}{ Acc. }\\
\midrule
COMET-22& $1$ & $86.90$& $2$ & $53.11$\\
BLCOM\_1& $1$ & $86.12$& $2$ & $53.00$\\
XCOMET& $1$ & $84.88$& $3$ & $52.35$\\
metametrics\_mt\_mqm\_hybrid\_kendall& $1$ & $84.80$& $1$ & $53.92$\\
XCOMET-QE*& $1$ & $83.67$& $4$ & $51.01$\\
PrismRefMedium& $1$ & $83.07$& $5$ & $50.62$\\
MetricX-24-Hybrid& $2$ & $82.96$& $2$ & $53.13$\\
BLEURT-20& $2$ & $82.19$& $2$ & $52.93$\\
gemba\_esa*& $2$ & $81.37$& $10$ & $40.86$\\
MetricX-24-Hybrid-QE*& $2$ & $80.93$& $4$ & $51.35$\\
PrismRefSmall& $2$ & $80.84$& $4$ & $51.23$\\
\cellcolor{highlight}MQM-2024& \cellcolor{highlight} $2$ & \cellcolor{highlight} $80.13$& \cellcolor{highlight} $11$ & \cellcolor{highlight} $34.61$\\
sentinel-cand-mqm*& $2$ & $79.06$& $6$ & $50.23$\\
YiSi-1& $2$ & $78.76$& $3$ & $51.88$\\
BERTScore& $2$ & $78.28$& $4$ & $50.88$\\
metametrics\_mt\_mqm\_qe\_kendall.seg.s*& $3$ & $77.05$& $7$ & $49.17$\\
CometKiwi-XXL*& $3$ & $76.51$& $5$ & $50.86$\\
bright-qe*& $3$ & $75.73$& $9$ & $43.46$\\
MEE4& $3$ & $75.20$& $4$ & $51.03$\\
CometKiwi*& $3$ & $74.39$& $5$ & $50.47$\\
chrfS& $3$ & $73.90$& $4$ & $51.05$\\
chrF& $4$ & $70.94$& $5$ & $50.69$\\
spBLEU& $4$ & $70.77$& $6$ & $49.81$\\
BLEU& $4$ & $69.37$& $7$ & $49.43$\\
damonmonli& $4$ & $63.34$& $8$ & $48.14$\\
sentinel-ref-mqm& $5$ & $54.36$& $12$ & $15.38$\\
sentinel-src-mqm*& $5$ & $54.36$& $12$ & $15.38$\\
XLsimMqm*& $5$ & $39.32$& $9$ & $43.02$\\
\bottomrule
\end{tabular}

    \caption{The test set is 2024. The evaluator selected as the ground truth follows the ESA protocol (ESA in Table~\ref{tab:full-wmt24}).}
    \label{tab:wmt24-gold-esa}
\end{table*}